\documentclass{article}

\usepackage[preprint]{neurips_2019}

\usepackage[utf8]{inputenc} 
\usepackage[T1]{fontenc}    
\usepackage{hyperref}       
\usepackage{url}            
\usepackage{booktabs}       
\usepackage{amsfonts}       
\usepackage{nicefrac}       
\usepackage{microtype}      

\usepackage{subfigure}
\usepackage{graphicx}
\usepackage{amsmath}
\usepackage{bm}
\usepackage{url}
\usepackage{stmaryrd}
\usepackage{balance}
\usepackage{amssymb}
\usepackage{pgfplots}
\usepackage{pifont}
\usepackage{epsfig}
\usepackage{booktabs}
\usepackage{pgffor}
\usepackage{tikz}
\usepackage{multirow}
\usepackage{amsmath}
\usepackage{autobreak}
\usepackage{wrapfig,lipsum}

\makeatletter
\newcommand\xleftrightarrow[2][]{%
	\ext@arrow 9999{\longleftrightarrowfill@}{#1}{#2}}
\newcommand\longleftrightarrowfill@{%
	\arrowfill@\leftarrow\relbar\rightarrow}
\makeatother

\newcommand{\CNN}{\textsc{CNN}}
\newcommand{\GNN}{\textsc{GNN}}
\newcommand{\bestm}{\textsc{GIN}}
\newcommand{\GGR}{\textsc{GGRNet}}
\newcommand{\GAIN}{\textsc{GAIN}}

\usepackage{algorithm}
\usepackage{algorithmic}

\begin{document}

\title{Multitask Learning On Graph Neural Networks Applied To Molecular Property Predictions}

\author{Fabio~Capela \\
	Firmenich SA \\ 
	CH-1227 Les Acacias, Geneva\\
	\texttt{fabio.andre.capela@firmenich.com}	 \\
	\And
	Vincent~Nouchi \\
	Firmenich SA \\ 
	CH-1227 Les Acacias, Geneva\\
	\texttt{vincent.nouchi@firmenich.com}	\\
	\And
	Ruud~Van~Deursen \\
	Firmenich SA\\ 
	CH-1227 Les Acacias, Geneva\\
	\texttt{ruud.van.deursen@firmenich.com}	\\
	\And
	Igor~V.~Tetko \\
	Helmholtz Zentrum München \\
	and BIGCHEM GmbH \\
	\texttt{itetko@bigchem.de}	\\
	\And
	Guillaume~Godin \\
	Firmenich SA\\ 
	CH-1227 Les Acacias, Geneva\\
	\texttt{guillaume.godin@firmenich.com}	\\
}

\maketitle

\begin{abstract}
	\vspace{-5pt}
	Prediction of molecular properties, including physico-chemical properties, is a challenging task in chemistry. Herein we present a new state-of-the-art multitask prediction method based on existing graph neural network models. We have used different architectures for our models and the results clearly demonstrate that multitask learning can improve model performance. Additionally, a significant reduction of variance in the models has been observed. Most importantly, datasets with a small amount of data points reach better results without the need of augmentation. 
\end{abstract}



\vspace*{-15pt}
\section{Introduction}\label{sec:introduction}
\vspace*{-10pt}
    
    Deep learning techniques typically require a large number of examples to be trained with as a result of the number of parameters used in the network architecture. For some applications, obtaining a large amount of data points is a challenging task. Measurements for some tasks can be considerably time-consuming or very expensive. One such group of measurements are physico-chemical properties of molecules. Despite the low availability of data, there is a strong need to develop highly predictive models for such properties. One such application includes the selection of molecules obtained from deep generative models. These deep generative models can generate millions of novel molecules in a short time and have become more powerful than enumeration methods\cite{JLRcopier, ruudpapergen, RuudGraph, JLRreview}. In summary, it is essential to obtain fast and accurate predicting models to identify new interesting molecules, e.g. as drug candidate or a new ingredient for perfumery industries\cite{sanchezlengeling2019machine}. In this article we investigate how these goals can be simultaneously achieved with the help of multitask learning using Graph Neural Networks ({\GNN}). 

    Chemical graph theory \cite{Diudea2018} represents molecules as undirected graphs using the vertices and edges for atoms and bonds, respectively. Recent advancements of deep learning methods for graphs, including {\GNN}, define an ideal setup to learn molecular properties, typically seen in classification and prediction tasks\cite{pathaugmentedgraph,kipf2016semi}. The advantage of using a graph as molecular representation is the replacement of the partial derivatives of a typical convolutional neural network ({\CNN}) by the Laplacian of the graph, which encodes in a natural way the topological features of a molecule. 
    
    {\GNN} is a powerful technique that has obtained state-of-the-art results in property predictions \cite{ggrnetpaper,3DGCN}. Even though it has been proven to be fairly easy to train {\GNN}s \cite{GNNfirst}, the previous studies typically learn one {\GNN} per property. Training individual neural networks for the dozens of physico-chemical properties could quickly turn into a tedious task. Moreover, the prediction time increases with the number of predictors, which is not practical when we would like to have an automatic screening of a large number of molecules.  
    
    One way to address the reduction of prediction times and possibly also to increase the accuracy of models is to consider multitask learning (MTL) in the context of {\GNN}. As first proposed by Caruana \cite{Caruana1997}, datasets with single targets have specific noise characterizations. Using MTL can lead to a more general model that is less dependent on specific noise patterns of individual datasets. Moreover, MTL has been proven \cite{Ruder2017AnOO} to solve the problem of inductive bias, i.e. choose a good hypothesis space, which is sufficiently generic to be able to learn within an environment of related tasks. An example includes the learning of related chemical properties such as solubility (LogS) and octanol-water partition  (LogP) or octanol-water distribution (logD) coefficients. MTL is expected to perform well on novel targets that have common ground with the tasks that the model was trained on. This is closely related to the problem of overfitting, which is dampened in the context of MTL as it implicitly acts as a regularization method. Moreover, MTL is a technique that naturally applies to {\GNN} as different chemical properties are frequently embedded in the molecular topology.


    \textbf{Graph Neural Networks for molecule property prediction} {\GNN}s have attracted considerable attention recently. It has been applied to different kinds of datasets that could be represented as a graph structure. Consequently, a plethora of architectures has been presented until now. Gilmer et al was the first to propose the message passing neural network \cite{MPNNarticle}, a category to which many modern {\GNN}s belong to.   
    
    Among the most interesting recent architectures, Keyulu Xu et al have proposed a graph isomorphism network (GIN) \cite{ginpaper}. The latter method is one of the most expressive methods among the large class of {\GNN}s, applying a neighborhood aggregation schemes at the node feature level. The essential message of the model is that the best way to update the node features at the k-th layer is through the usage of
    \begin{equation}
    \label{eqn:ginconv}
        h_v^{(k)} = \text{MLP}^{(k)}\left(\left(1+\epsilon^{(k)} \right)\cdot h_{v}^{(k-1)}+ \sum_{u\in \mathcal{N}(v)} h_{u}^{(k-1)} \right)
    \end{equation}
    where $h_{v}^{(k)}$ represents the features at node $v$ for the $k$-th layer. $\text{MLP}$ is a multi-layer perceptron given by two fully connected layers with a ReLU activation function on the first layer. $\epsilon^{(k)}$ might be a learnable parameter or simply be put to 0 (GIN-0).$\mathcal{N}(v)$ is the neighborhood of a node $v$ in a particular graph. The GIN model was applied to several classification tasks, such as MUTAG dataset, PROTEINS dataset, where it reaches state-of-the-art results \cite{ginpaper}. 
    
    There are also models that take into account global properties of the molecule on top of using local neighborhood aggregation schemes. Shindo et al. in \cite{ggrnetpaper} introduced two features on top of the atom feature embeddings: a counting feature $x_N$ and vertices distance feature $d_{v,u}$. The counting feature corresponds to the total number of atoms on a molecule, while the distance feature is the reciprocal of the Euclidean distance between each pair of nodes. Such features give global information to the {\GNN}s about the structure of the molecule. Apart from the extra features, they also introduced a gated graph recursive neural network (GGRNet). GGRNet uses the idea of skip connections and shared information throughout the multiple layers of the graph, mimicking residual neural networks for {\GNN}s.
    
    Another interesting idea in \cite{gatpaper} is the introduction of an attention mechanism applied to {\GNN}, the so-called GAT model. Such model allows to learn the most important features among a neighborhood of nodes, yielding a learned mask. In particular, they learn the attention coefficients $e_{ij}$ for $j \in \mathcal{N}(i)$. Then, such coefficients are normalized to be comparable across the overall nodes through 
    \begin{equation*}
        \alpha_{ij} = \text{softmax}_{j}(e_{ij}) = \frac{\exp(e_{ij})}{\sum_{k\in \mathcal{N}_i} \exp(e_{ik})}.
    \end{equation*}
    
    In this work we have decided to explore different GNN architectures for MTL learning.
    
    \textbf{MTL applied to molecular properties} A review of MTL techniques used in chemistry has been presented in \cite{IgorMTL}. One of the first studies on the use of MTL in chemistry was done by one of the  authors of this study. It was demonstrated that simultaneous prediction of eleven tissue-air partitioning coefficients using neural networks had significantly lower errors compared to that of individual properties.~\cite{varnek2009}. Three recent works are also interesting to report. The first one is the work from Dahl et al.~\cite{dahl2014multi}, in which they explored how effective multitask neural networks are for QSAR applications. They fitted 19 targets through a multi-target deep neural network (MT-DNN), where the dataset was 100'000 molecules coming from PubChem with 3'764 descriptors per molecule generated by Dragon. Their results are striking: MTL outperformed all of the previous state-of-the-art models for 14 out of 19 targets and showed comparable results for the other 5 targets. They reported that MTL-DNN produced a more generic embedding that can be applied to other tasks. Moreover, the weights, being trained on more data, lead to less variable results as shown by reduced standard deviations. Interestingly, they have also concluded that the model was indifferent to highly correlated features, which implies that no degradation of results has been observed when halving the number of input features.

    A second work on MTL-DNN was provided by Hochreiter and co-workers in \cite{HochreiterDNN}. In this work, the full ChEMBL database has been used. It contains 743,336 molecules with 13 millions sparse features for each compound and provided benchmarks for >1'000 biomolecular targets. The MTL-DNN model outperformed all the standard machine learning models with AUCs > 0.8. The authors explained that the results obtained were due to the shared hidden representation that led to more conservative results addressing the problem of acute outliers in predictions.
    
    In the context of {\GNN}, MTL has already been discussed in \cite{fare2018powerful}. In this work, the authors have built so-called MT-Weave graph architectures with or without transfer learning, applied to 48 tasks including regression and classification. The architectures fall under the framework of message passing neural networks and {\GNN}, where atom and pair features on every layer are passed to the next layer, very much like convolutional neural networks. 
    
    Overall, previous works concluded that MTL applied to property prediction have mainly two advantages: (i) multitask models utilize relations among the tasks that produce overall better results than using single task models, (ii) shared hidden embeddings are more effective representations as some targets may have very few measurements.

    In the present work, we study the effects of MTL on graph neural networks applied to the prediction of chemical properties at the molecular level. We train different types of graph neural networks on tasks that are related and therefore share a common hypothesis space. We make a study of the different synergistic effects of different datasets. In particular, we discuss how to deal with datasets that have missing targets for specific molecules. 
        The main contributions of the present paper are: 
        
        \begin{itemize}
            \item Study of MTL applied to graph neural networks with convolutions, attention and gated layers
            \item Automatic selection of targets that have a synergistic effect
            \item Selection study of dataset combinations to obtain the best results 
            \item Impact analysis of multitask based on the size of the datasets 
            \item Study of the explanation of contributing effect on multitask learning 
        \end{itemize}

    Additionally, we also leverage the learned weights of generalized tasks to other specific tasks by retraining the weights of the last layer, while freezing the weights of all other layers. Lastly, we have also evaluated the time of inference for both MTL and sequential inference on single tasks.

\vspace*{-10pt}
\section{Terminology and Problem Definition} \label{sec:formulation}
\vspace*{-5pt}

In this section, we define the terms and notations used in the present paper and provide a formulation for the MTL on {\GNN}.

\vspace*{-10pt}
\subsection{Problem Formulation}
\vspace*{-5pt}

    A graph is represented by $G=(\mathcal{V},\mathcal{E})$, where node feature vectors $X_{v}$ for $v \in \mathcal{V}$. {\GNN}s follow certain aggregation strategies to learn the representation of a node or of an entire graph. In our particular case, we would like to learn graph regression, using a set of initial graphs $\{G_1,G_2, \dots,G_n\}=\mathcal{G}$, representing the molecules and their labels $\{\mathcal{Y}_1, \dots,\mathcal{Y}_k\}=\mathcal{Y}$, corresponding to the $k$ molecular properties.


\vspace*{-8pt}
\section{Proposed Models}\label{sec:models}
\vspace*{-8pt}

    In this section, we introduce our proposed models in detail. We present the overall architectures and provide details about some of the modifications introduced into the existing architectures. Our experiments are based on three modified graph neural networks: GGRNet, a graph attention isomorphism network (GAIN) and GIN. These three topologies are shown in Figure (1). The implementations are coded using pytorch geometric library\cite{Fey/Lenssen/2019}. 
    
    \textbf{{\GGR}} In the following, we have considered the model of {\GGR} with only the counting feature $x_N$, not including the distance feature $d_{v,u}$ discussed in the paper \cite{ggrnetpaper}. Overall, we used 47 one-hot encoded features and $x_N$ for all the models considered in the paper. Then, we have also added in the first layer of the network a graph isomorphism convolution from eq. (\ref{eqn:ginconv}) with a ReLU activation function, followed by a batch normalization. Our initial node feature representation $h_{v}^0$, instead of $h_v^0=0$ for all $v$ as explained in the \cite{ggrnetpaper}, goes as
    \begin{equation*}
        h_v^0 = \text{BN}\left\{\text{ReLU}\left[\text{MLP}\left(x_v+\sum_{u \in \mathcal{N}(v)}x_u\right)\right]\right\}.
    \end{equation*}
    The MLP considered is composed of two fully connected layers with a ReLU function after the first layer. This initialization is followed by the gated recursion explained in the Ref.~\cite{ggrnetpaper}, where we did 10 iterations. The final part of the graph network is a MLP with the same architecture of the initialization part with 95 neurons in the hidden layer and the number of targets in the output.
    
    \textbf{GAIN} This model borrows the attention mechanism described previously in \cite{gatpaper}. We used such mechanism as a self-learned mask that is able to select the features, among a neighborhood of nodes, that are important to pass to the next layer. Apart from the self-attention layer, we also used a graph isomorphism convolution and two fully connected layers.
    
    \textbf{GIN} The GIN model served as the basis for the three models created in the present paper. We reduced the complexity of the original model \cite{ginpaper} by considering only one convolutional layer. We also added two fully connected layers at the end, like for {\GGR} and GAIN.
    
    \begin{figure}
        \center
        \includegraphics[width=0.9\linewidth]{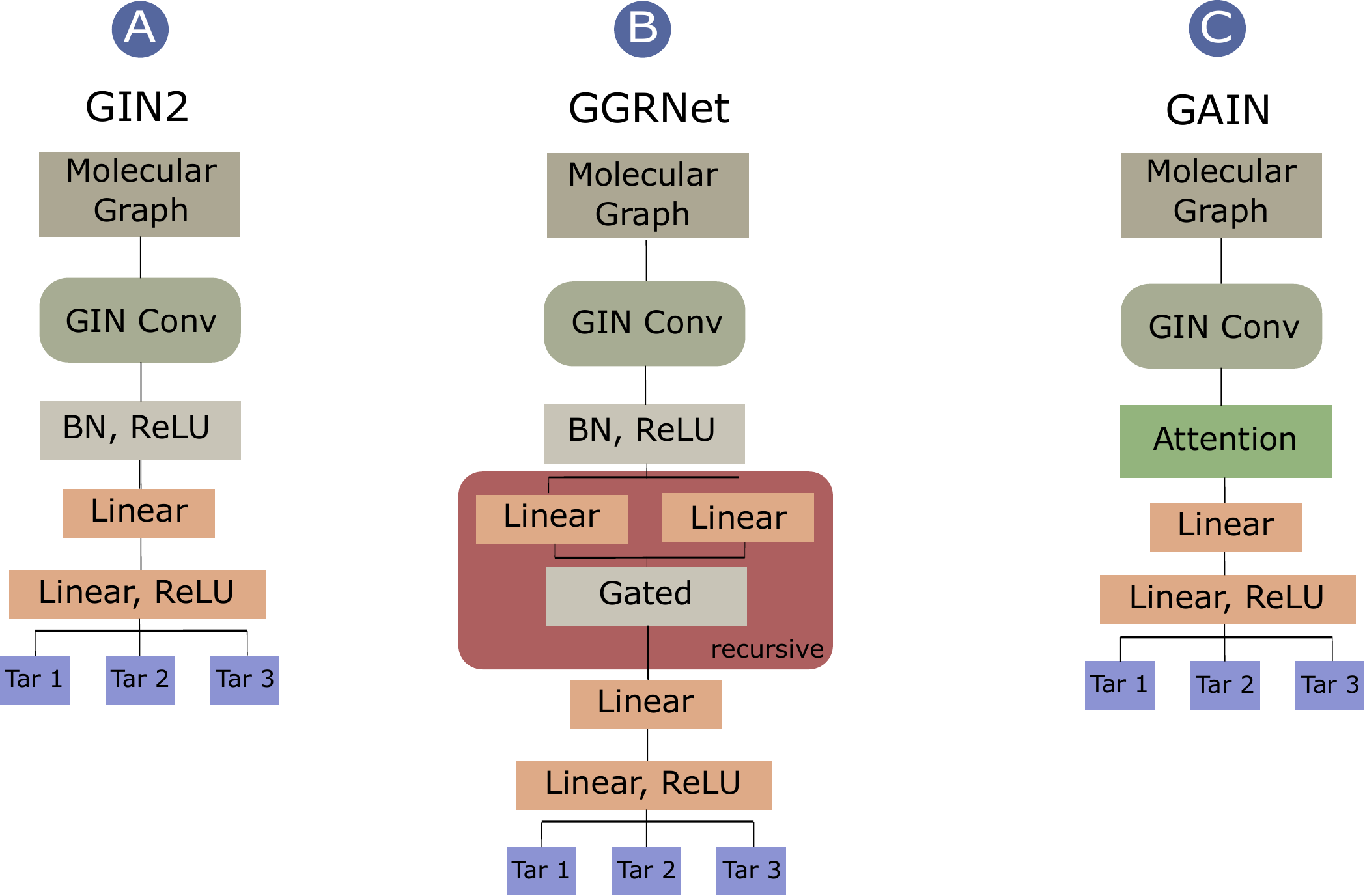}
        \caption{Architecture of the three models used throughout the paper: GIN (A), {\GGR} (B) and GAIN (C). Different layers are used  as the central ingredients in different models. The graph isomorphism convolution and the two fully connected layers at the bottom are maintained on all the models. The number of targets are variable. If we restrict the models to one target in the last layer, the architecture is open to single task learning.}\label{fig:1}
    \end{figure}

\vspace*{-10pt}
\section{Results}\label{sec:experiment}
\vspace*{-8pt}

    To evaluate the performances of MTL on the different models, we have used a set of physico-chemical properties. We discuss their selection process for the MTL model. The main point investigated is whether MTL gets better results than single task models for different chemical properties. We also briefly focus on the impact of the size of the dataset in multitask learning. The size is connected to an implicit data augmentation effect. We also evaluated the time of inference for both MTL models and single task models. Finally, we examined the effect of transferability of the latent space learned in the isomorphism convolution through a transfer learning setup.

\subsection{Datasets and experimental settings}
\vspace*{-5pt}                         

\begin{wraptable}{l}{5.5cm}
\caption{Datasets}\label{wrap-tab:1}
\center
\begin{tabular}{cc}\\\toprule  
Dataset & \# Molecules \\\midrule
Esol & 1'128 \\  
FreeSolv &613* \\  
logD7.4 &4'200 \\  
Boiling point &5'435 \\  
LogVP &2'708 \\  
LogP &13'973 \\  

\bottomrule
\end{tabular}
\end{wraptable} 

     In our experiments we used several public datasets for regression tasks. The set of evaluated properties include ESOL \cite{esolarticle}, FreeSolv \cite{freesolvarticle}, logD7.4  \cite{lipodataset}, boiling point (BP), vapor pressure (LogVP) and octanol-water partition coefficient (logP)\cite{Sushko2011}. The datasets define different physico-chemical properties of molecules and are summarized (Tab.~\ref{wrap-tab:1}). For example, logD7.4 corresponds to the experimental results of octanol-water distribution coefficient at pH=7.4 of 4'200 compounds. ESOL has 1'128 compounds with the experimental values of water solubility. FreeSolv dataset contains experimental values for the hydration free energy of small molecules. FreeSolv contains 643 molecules and is also the smallest evaluated dataset in this work. 
    
     \begin{figure}
        \centering
            \subfigure[Histogram of FreeSolv]{\includegraphics[width=0.48\textwidth]{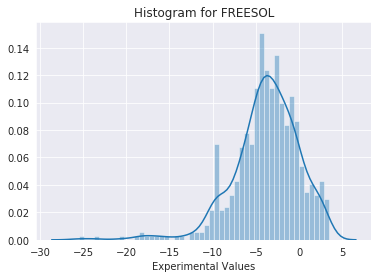}}
            \subfigure[BP predictions with multitask GIN model]{\includegraphics[width=0.48\textwidth]{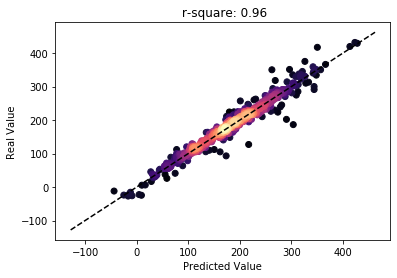}}
        \caption{(a) The normalized count of the experimental values of FreeSolv, showing a long tail of points on the left. (b) A scatter plot representing the predictions from a GIN MTL model vs the experimental values of the boling point (BP) dataset with a correlation coefficient ($R^2$) of 0.96}\label{fig:2}
    \end{figure}

\begin{wrapfigure}{r}{0.4\textwidth}
        \centering
        \includegraphics[width=0.38\textwidth]{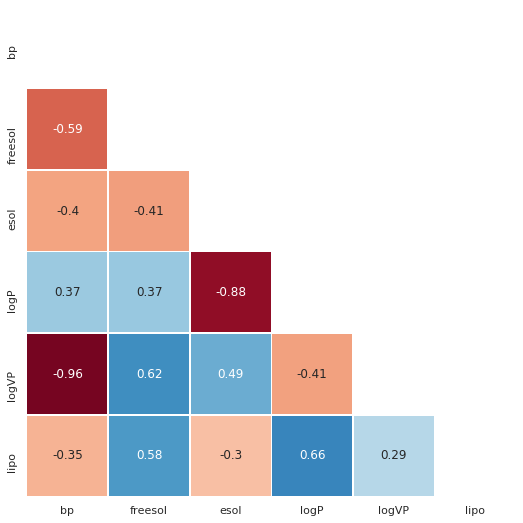}
        \caption{Cross-correlation matrix among the datasets.}\label{wrap-fig:1}
    \end{wrapfigure}

     A cross-correlation matrix is represented in Fig.\ref{wrap-fig:1}. We trained the models with datasets that have the highest correlation, i.e. logD7.4/LogP, Esol/LogP and FreeSolv/LogVP/BP. We obtained better results for related tasks than tasks with lower correlation (see Tab~\ref{tab:data_selection} and appendix \ref{appendix:data_selection}). 
     
\subsection{Automatic Target Selection}
\vspace*{-5pt}  
     We have implemented a script that looks at the correlation between pair of targets and automatically selects the pair of properties that are highly correlated in absolute value. Then, it constructs a list that have at least one common element among the pairs. We have used a threshold correlation of 0.5. Such script can be used when the list of targets is very large. In our particular case, we have decided to split logD7.4, Esol and LogP into two buckets, since it provided slightly better results.
      
\subsection{FreeSolv Data Selection For Fair Comparison}
\vspace*{-5pt}  
    For the FreeSolv dataset, there is a considerable spread in results (RMSE) among models in the literature. For example, \cite{gaanpaper} reported a RMSE of $0.294\pm0.005$, while Hyeoncheol Cho et al. in \cite{3DGCN} has a RMSE of  $0.828 \pm 0.126$. Several other publications report wildly changing results \cite{geometricgcn}. All of these techniques are very similar in essence and should therefore display similar results.The wild variations in the results might be due to the 2 following reasons: FreeSolv is a small dataset of 643 molecules and it contains a very sparse flat tail (see Fig.~\ref{fig:2}). Therefore, a simple split between the train, validation and test set following a 8:1:1 ratio intrinsically leads to very diverse results. Indeed, in some cases the test set might contain a high percentage of outliers, while in other cases it does not contain a single outlier. Following this observation, we excluded molecules with values < -10 from the dataset, eliminating 29 molecules from evaluation. Additionally, all results presented for FreeSolv were obtained averaging 4 different splits with different seeds. 
    
\subsection{Setup and Evaluation Metrics}
\vspace*{-5pt}     
    
    In our experiments, we have partitioned the train/test according to 8:2 split. We then performed a 5-fold cross-validation on the training set to evaluate the reproducibility of the used architecture. At the end of each fold, we measured the loss on the respective test set. The batch size used for our experiments was 300. The learning rate was initialized at 0.01, applying a learning rate scheduler with $\gamma=0.5$. This scheduler multiplied the learning rate by $\gamma$, if the validation loss did not decrease after 40 epochs. Each dataset was trained separately and in conjunction with other targets as previously explained. The dimension of the hidden layer was 95 neurons, which is an optimized parameter. Moreover, for {\GGR} we have considered 10 recursions. For the GAIN model, we kept one head attention with a dropout of 30\% that is applied to the normalized attention coefficients. Most of the parameters were not optimized for the trained tasks, but kept the same between single task training and MTL for comparison purposes.  

    In the present paper, we deal with sparse datasets, meaning that not all molecules displayed values for all evaluated targets. Moreover, a large set of molecules typically display a value for a single end point. To backpropagate the error correctly, we applied a modified cost function addressing the sparsity in the dataset:
    \begin{equation}
    \mathcal{L}_{\text{tot}} = \left(\frac{1}{n}\sum_{k}^K\sum_{i-1}^{n} \delta_{kj} \left(\mathcal{Y}_k^{i}-\hat{\mathcal{Y}_k}^{i}\right)^2 \right)^{1/2},  
    \label{eqn:ourloss}
    \end{equation}

    where the $k$ index is related to the target under consideration. The formula \ref{eqn:ourloss} reduces to the root mean squared error (RMSE) in the case of a single task. To dismiss the values that do not have a target, we applied a mask on a per batch basis and backpropagate the error only on the relevant instances that have at least one target present. Predictions on the test set were performed using the full length of the set. The results (RMSE) for multitask learning and single task learning for all evaluated architectures are summarized in table 2.

\vspace*{-10pt}
\subsection{Experimental Results}
\vspace*{-5pt}

    The results of our experiments are shown in Table \ref{tab:classification_result}. For all datasets except LogP, MTL showed improved results for the overall RMSE. MTL did no show an improvement for LogP (shown in italic) and we believe this may be an effect of the dataset size. Indeed, LogP defined the largest dataset with 13'973 molecules. The benefit of MTL for larger datasets is relatively small, in particular in datasets that are sufficiently large to produce highly reliable models in single task learning. The effect of the size was evaluated varying the size of the LogP-dataset ~\ref{fig:4}. To create such plot, we have varied the size of the training LogP data from 20\% to 90\% of the full set, while having the same test set throughout the process. Then, we have calculated the percentage improvement that MTL brings to the overall RMSE on the test set. The results show that MTL is an essential benefit to improve the performance on smaller datasets, because it provides a larger and more diverse set of molecules for the network to learn from and can compensate for the limited amount of data. Contrarily, for large datasets, the typical MTL improvement becomes negligeable or inexistent. We consider this is an effect of implicit data augmentation in multitask learning.  Also, the effect of MTL is perceivable for small datasets when they are put together with other datasets that are larger and share a common hypothesis space. 
    
    MTL improves the results by ~4\%-20\% in comparison to single task models. In bold in Tab.~\ref{tab:classification_result}, we have shown the values that are statistically better from their single task counterpart. The performances vary between used architectures. {\GGR} displays the best results for Esol, FreeSolv and LogVP. GIN obtainss the lowest RMSE for logD7.4, boiling point and logP. Finally, GAIN shows considerable improvements in comparison to single task learning. In particular, the improvements for FreeSolv and BP stand out. We believe that the dropout applied to the attention mechanism may affect the results in single task models for small datasets, such as FreeSolv. Indeed, for small datasets models are typically in underfitting regime. MTL, with its implicit data augmentation effect can adequately address this regime and yield considerable improvements on small datasets.

\begin{table*}[t]
	\vspace*{-20pt}
	\caption{Results (RMSE) for MTL and single task on different graph models.}
	\label{tab:classification_result}
	\centering
	{
		\begin{tabular}{l ccc ccc}
			\toprule
			&\multicolumn{3}{c}{Multitask Models} & \multicolumn{3}{c}{Single Task Models}\\
			\cmidrule(lr){2-4} \cmidrule(lr){5-7}
			Dataset  &{\GGR} &{\GAIN}  &{\bestm} &{\GGR} &{\GAIN}  &{\bestm}  \\
			\midrule
			Esol 
			&$\mathbf{0.53 \pm 0.02}$ 
			&$0.62 \pm 0.02$ 
			&$\mathbf{0.59 \pm 0.03}$ 
			&$0.65 \pm 0.03$ 
			&$0.63 \pm 0.04$ 
			&$0.65 \pm 0.03$ \\ 
			
			FreeSolv 
			&$0.80 \pm 0.03$ 
			&$\mathbf{0.87 \pm 0.03}$ 
			&$0.83 \pm 0.03$ 
			&$0.82 \pm 0.05$ 
			&$1.04 \pm 0.06$ 
			&$0.89 \pm 0.04$ \\ 
			
			logD7.4 
			&$0.69 \pm 0.02$ 
			&$0.71 \pm 0.05$ 
			&$\mathbf{0.64 \pm 0.02}$ 
			&$0.74 \pm 0.03$ 
			&$0.72 \pm 0.05$ 
			&$0.74 \pm 0.03$ \\ 
			
			BP 
			&$18.9 \pm 0.5$ 
			&$18.8 \pm 0.7$ 
			&$\mathbf{18.0 \pm 0.2}$ 
			&$19.9 \pm 1.9$ 
			&$19.6 \pm 0.8$ 
			&$22.7 \pm 1.6$ \\ 
			
			LogVP 
			&$\mathbf{0.89 \pm 0.02}$ 
			&$0.93 \pm 0.04$ 
			&$0.98 \pm 0.04$ 
			&$0.94 \pm 0.02$ 
			&$0.96 \pm 0.04$ 
			&$0.99 \pm 0.04$ \\ 
			
            LogP 
            &$0.50 \pm 0.01$ 
            &$0.50 \pm 0.03$ 
            &$0.47 \pm 0.01$ 
            &$0.50 \pm 0.02$ 
            &$\mathit{0.45 \pm 0.02}$ 
            &$0.48 \pm 0.02$ \\ 
			\bottomrule
			
		\end{tabular}
	}
	\vspace*{-10pt}
\end{table*}

\vspace*{-5pt}
\subsection{Loss Convergence}
\vspace*{-5pt}

\begin{figure}
        \centering
            \subfigure[MTL Improvement as a function of size]{\includegraphics[width=0.48\textwidth]{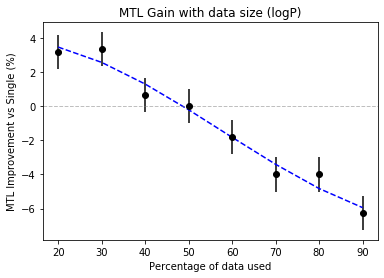}}
            \subfigure[Loss decrease for both MTL and single model]{\includegraphics[width=0.48\textwidth]{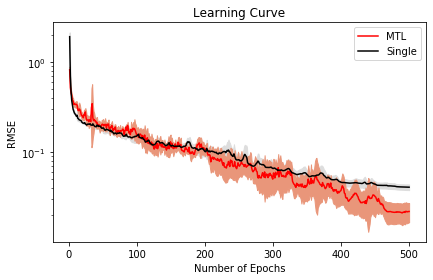}}
        \caption{(a) The results for the effect of MTL on the size of the dataset (logP). As it is clear from the figure, the positive effect of MTL decreases with increasing size. (b) Comparison of the loss in single task and multitask learning using the GIN-architecture, showing that the loss in multitask learning descends more rapidly.}\label{fig:4}
    \end{figure}

    On the right side of the Fig.~\ref{fig:4}, we can see that the loss decreases for both MTL and single task learning on a particular task (BP for the case of the figure). MTL reaches much lower RMSE for the same number of epochs as the single task learning. Indeed, MTL achieves levels of the loss 150 epochs before single task arrives at similar values. This is mainly due to the effect of data augmentation that is present in MTL. 
    
\vspace*{-5pt}
\subsection{Weights and Constrained System}
\vspace*{-5pt}

    To understand why MTL provides typically better results than single task, we have evaluated the weights of the fully connected hidden layer, directly prior to the output layer. The comparison for multitask learning and single task learning has been summarized in Fig.~\ref{fig:5}. We observe that the weights in the single task define very narrow distributions, while the spread is significantly larger for MTL. A requirement of networks to have smaller weights (which depending on the used normalization is equivalent to their wider distribution) in order to provide better generalization was theoretically justified by \cite{SEPP}. We hypothesize that low spread of weights may be an effect of models getting trapped in local minimums for single task learning, whereas MTL is required to search for a better balance in the weights to address multiple tasks simultaneously. Thus, MTL provides a regularisation of the weights by making them more equally distributed. This reduces the risk for a model to get trapped in a local minimum and contribute to a better performance of models. Such behaviour may, however, not always be the case and could depend on the size of the datasets. Datasets with large sizes more significantly influence the training process, while smaller datasets have a reduced influence on the training process. These datasets are, however, the primary beneficiaries in multitask learning as a result of implicit data augmentation.

\vspace*{-5pt}
\subsection{Inference Time}
\vspace*{-5pt}

    \begin{wrapfigure}{l}{0.4\textwidth}
        \centering
        \includegraphics[width=0.38\textwidth]{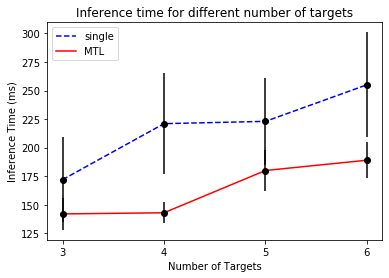}
        \caption{Time inference study for single and multitask models.}\label{wrapfig:2}
    \end{wrapfigure}

    We have also evaluated the inference time needed for multitask and single task predictions. We trained independently a single model for each task and the MTL models from 3 to 6 tasks. We used GIN as the base model. The inference time was measured using a CPU on 648 molecules. The results are summarized in Fig.~\ref{wrapfig:2}. The results show that MTL is typically 25\% faster than a single task process. The difference in inference time increases with number of tasks. We observe values of 33\% and 48\% for multitask and single task models, respectively. Thus, in addition to improving model performance, MTL is also beneficial for inference time in comparison to single task models.

\vspace*{-5pt}
\subsection{Transfer Learning}
\vspace*{-5pt}

    \begin{figure}
    
        \centering
            \subfigure[Histogram of the weights for the single task model]{\includegraphics[width=0.48\textwidth]{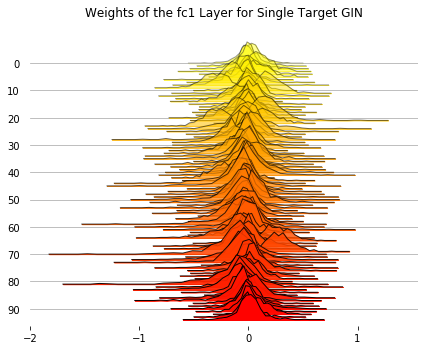}}
            \subfigure[Histogram of the weights for the multitask model]{\includegraphics[width=0.48\textwidth]{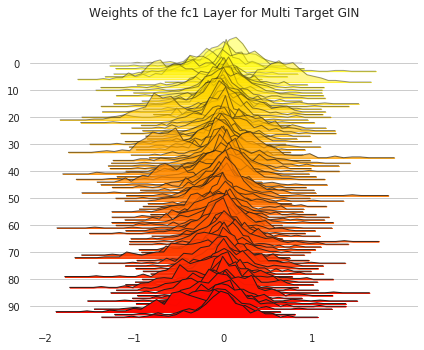}}
        \caption{Matrix of the weights of one fully connected layer for (a) single task model and (b) multitask models.}\label{fig:5}
        
    \end{figure}

    In transfer learning, one typically poses the question whether a set of general features can be learned, which are subsequently applied to a new task. In transfer learning, the top layers typically have features that contain general information about the problem and that can be applied to many other tasks. To understand the transferability of the top graph isomorphism convolutional layer to tasks not included in the training set, we isolated one task and trained the model with all remaining tasks. The used architecture is GIN. After training of the multitask model, we have subsequently transferred the weights of the top layers to a new model. The model was then retrained using the isolated task, only updating the weights for the two last fully connected layers. The top layers with transferred weights were not updated during this training step.
    The results of transfer learning show that there is no improvement compared to single task learning, i.e. no significant reduction of RMSE (see Tab.~\ref{tab:transfer_learning}). These results suggest that the transferability in these graph neural networks is limited. Moreover, some tasks even displayed negative effects on transfer learning. The number of tasks might play a role on transferability, as explained in \cite{massivemultitaskpaper}. These results suggests that the primary reason for the improvement may be due to an implicit data augmentation obtained from merging the datasets for different but similar end points.

\vspace{-10pt}
\section{Conclusion}\label{sec:conclusion}
\vspace{-10pt}

Herein we have evaluated the effect of MTL on graph neural networks for three types of architectures, i.e. GIN, GGRNet and GAIN. The architectures of these networks vary slightly and contain different central elements such as an attention layer, a graph isomorphism convolution and a recursive neural network with a gating function. The results obtained with the multitask models are consistent across studied architectures and on average are similar or frequently significantly better than the results of single task models using the same architectures. In particular, smaller datasets benefit from multitask learning. Indeed, the results for the largest dataset, LogP, were the least influenced by the presence of the other tasks. We have also evaluated the benefit of multitask learning while increasing the size of the datasets, producing better results for smaller size datasets. This effect seems to be primarily a result of an implicit data augmentation when merging datasets for multitask learning, which is further supported by the negligible transferability observed using transfer learning. This conclusion is in line with earlier reported results\cite{massivemultitaskpaper}. Additionally, we evaluated the evolution of weights of GNNs throughout the training. While single task models display narrow distributions, multitask models keep wider distributions. We believe that this effect is essential to find the best compromise between the tasks in the training set. 

\textit{Note added}. At the final stage of this project, we became aware of the related work carried out by the Bayer team \cite{bayermtlgnn}. Even though the end results are similar, the main differences are (i) the datasets used, (ii) the graph architectures and (iii) the preselection of the targets. In addition, we have investigated the transferability of the embedding coming from the convolutional layer (transfer learning), conducted a rigorous experiment on the size effect of the datasets on the results and studied the distribution of the weights for both single and multitask models. Moreover, our code and datasets are available on github (\href{firmenich/MultiTask-GNN}{https://github.com/firmenich/MultiTask-GNN}).

\vspace{-10pt}
\section*{Declarations}\label{sec:Declarations}
\subsection*{Availability of data and materials}
The code and used datasets are available on gitbub under a BSD-3 license: \href{https://github.com/firmenich/MultiTask-GNN}{https://github.com/firmenich/MultiTask-GNN} as well as it is available online as part of the OCHEM (http://ochem.eu) platform.
\vspace*{-5pt}
\subsection*{Funding}
FC, GG, VN and RvD are full-time employees of Firmenich SA, Geneva, Switzerland and funded by internal company sources. IVT is funded by Helmholtz Zentrum München and BIGCHEM GmbH.
\vspace*{-5pt}
\subsection*{Acknowledgement}
The authors thank Sven Jeanrenaud of Firmenich SA, Geneva, Switzerland for critical reading of the document.
\vspace*{-5pt}
\subsection*{Competing interests}
The authors declare that they have no competing interests.
\vspace*{-5pt}
\subsection*{Contributions}
Architecture has been defined by FC, GG and IT. GG and RvD prepared and cleaned the individual datasets and defined the final merged dataset for multitask learning. FC coded all architectures. FC, VN and GG performed calculations. IT, GG and FC included architectures in OCHEM. All authors contributed equally to the writing of the manuscript.
\vspace*{10pt}

\pagebreak

\appendix
\section*{Appendix}
\subsection*{A. Results of MTL with no data selection} \label{appendix:data_selection}
\vspace*{5pt}
\begin{table*}[h!]
	\vspace*{-20pt}
	\caption{Results (RMSE) for MTL with data selection vs without data selection}
	\label{tab:data_selection}
	\centering
	{
		\begin{tabular}{l ccc ccc}
			\toprule
			&\multicolumn{3}{c}{Without Data Selection} &\multicolumn{3}{c}{With Data Selection} \\
			\cmidrule(lr){2-4} \cmidrule(lr){5-7}
			Dataset  & GIN & GGRNet & GAIN & GIN & GGRNet & GAIN\\
			\midrule
			Esol 
            &$0.61 \pm 0.08$ 
            &$0.62 \pm 0.03$ 
            &$0.62 \pm 0.04$ 
            
            &$\mathbf{0.59 \pm 0.03}$ 
            &$\mathbf{0.53 \pm 0.02}$ 
            &$\mathbf{0.62 \pm 0.02}$ \\

			FreeSolv 
			&$1.15 \pm 0.11$ 
			&$1.00 \pm 0.04$ 
			&$1.25 \pm 0.04$ 
            
            &$\mathbf{0.83 \pm 0.03}$ 
            &$\mathbf{0.80 \pm 0.03}$ 
            &$\mathbf{0.87 \pm 0.03}$\\ 

			logD7.4 
			&$0.64 \pm 0.03$ 
			&$\mathbf{0.68 \pm 0.04}$ 
			&$0.72 \pm 0.05$ 
			
			&$\mathbf{0.64 \pm 0.02}$ 
			&$0.69 \pm 0.02$ 
			&$\mathbf{0.71 \pm 0.02}$\\ 
			
			BP 
			&$18.4 \pm 0.3$ 
			&$19.2 \pm 1.1$ 
			&$18.9 \pm 0.4$ 
			
			&$\mathbf{18.0 \pm 0.2}$ 
			&$\mathbf{18.9 \pm 0.5}$ 
			&$\mathbf{18.8 \pm 0.7}$\\ 

			LogVP 
			&$\mathbf{0.96 \pm 0.05}$ 
			&$1.00 \pm 0.06$ 
			&$1.01 \pm 0.04$ 
			
			&$0.98 \pm 0.04$ 
			&$\mathbf{0.89 \pm 0.02}$ 
			&$\mathbf{0.93 \pm 0.04}$\\ 

            LogP 
            &$0.51 \pm 0.02$ 
            &$0.53 \pm 0.02$ 
            &$0.55 \pm 0.02$ 

            &$\mathbf{0.47 \pm 0.01}$ 
            &$\mathbf{0.50 \pm 0.01}$ 
            &$\mathbf{0.50 \pm 0.03}$ \\
            \bottomrule
			
		\end{tabular}
	}
	\vspace*{-10pt}
\end{table*}

To confirm that the dataset selection, based on the correlations as discussed in the main text, is beneficial to the reduction of the loss, we have run the MTL models on the 6 tasks discussed. The results shown in Tab.~\ref{tab:data_selection} demonstrate that a data selection provides better results (lower RMSE) compared to no selection of the final targets at all, apart for two datasets: logD7.4 and LogVP, which might be in a underfitting regime.

\subsection*{B. Results on Transfer Learning}
\vspace*{5pt}
\begin{table*}[h!]
	\vspace*{-20pt}
	\caption{Results (RMSE) for Transfer Learning on each task vs Single Models}
	\label{tab:transfer_learning}
	\centering
	{
		\begin{tabular}{l cc cc}
			\toprule
			Dataset  & GIN (Single) & GIN (Transfer Learning) \\
			\midrule
			Esol 
            &$0.65 \pm 0.03$ 
            &$0.76 \pm 0.04$ \\

			FreeSolv 
			&$0.89 \pm 0.04$ 
            &$0.85 \pm 0.01$ \\

			logD7.4 
			&$0.74 \pm 0.03$ 
			&$0.92 \pm 0.02$\\ 
			
			BP 
			&$22.7 \pm 1.6$ 
			&$21.6 \pm 0.92$\\ 
			
			LogVP 
			&$0.99 \pm 0.04$ 
			&$1.06 \pm 0.04$\\ 
			
            LogP 
            &$0.48 \pm 0.02$ 
            &$0.61 \pm 0.03$ \\
            \bottomrule
			
		\end{tabular}
	}
	\vspace*{-10pt}
\end{table*}

To test transfer learning, we have trained a full GIN model on 5 tasks and leave one task out. Then, we have freezed all the layers related to the convolutions, which are feature generators, and retrained only the fully connected layers on the task that was left out. The results are presented for the GIN model, both on single task and for transfer learning. As it is visible from the Tab.~ \ref{tab:transfer_learning}, we don't see any kind of transferability of the features trained by the graph isomorphism convolution to the task that was left out, apart from minor improvements for FreeSolv and BP that are not statistically significant.

\bibliographystyle{unsrt}
\bibliography{biblio}
\end{document}